\documentclass{article}

\usepackage{arxiv}

\usepackage[utf8]{inputenc} 
\usepackage[T1]{fontenc}    
\usepackage{hyperref}       
\usepackage{url}            
\usepackage{booktabs}       
\usepackage{amsfonts}       
\usepackage{nicefrac}       
\usepackage{enumerate}
\usepackage{microtype}      
\usepackage{graphicx}
\usepackage{amsmath}
\usepackage{threeparttable}
\newcommand{\norm}[1]{\left\lVert#1\right\rVert}

\title{A Pose Proposal and Refinement Network for Better 6D Object Pose Estimation}

\author{
      Ameni Trabelsi, Mohamed Chaabane, Nathaniel Blanchard, and Ross Beveridge  \\
  Department of Computer Science, Colorado State University }

\begin{document}
\maketitle

\begin{abstract}
In this paper, we present a novel, end-to-end 6D object pose estimation method that operates on RGB inputs. Our approach is composed of 2 main components:
the first component classifies the objects in the input image and proposes an initial 6D pose estimate through a multi-task, CNN-based encoder/multi-decoder module.
The second component, a refinement module, includes a renderer and a multi-attentional pose refinement network, which iteratively refines the estimated poses by utilizing both appearance features and flow vectors. Our refiner takes advantage of the hybrid representation of the initial pose estimates to predict the relative errors with respect to the target poses. It is further augmented by a spatial multi-attention block that emphasizes objects' discriminative feature parts. Experiments on three benchmarks for 6D pose estimation show that our proposed pipeline outperforms state-of-the-art RGB-based methods with competitive runtime performance. 

\end{abstract}

\section{Introduction}
Accurate 6D object pose estimation is crucial for many real-world applications, such as autonomous driving, robotic manipulation, and augmented reality. For instance, a 6D pose estimator for robot grasping needs to balance accuracy, robustness, and speed to be realistically deployable in real-world scenarios. 

Some approaches \cite{wang2019densefusion,xu2018pointfusion} have relied upon 
depth information in order to boost reliability and accuracy. However, depth sensors suffer a variety of failure cases, have high energy and monetary costs, and are less ubiquitous than their non-depth counterparts. Ultimately, pose estimation from RGB alone is a more challenging problem, but also a far more attractive option. 

\begin{figure}[t]
\centering
\includegraphics[height=7cm]{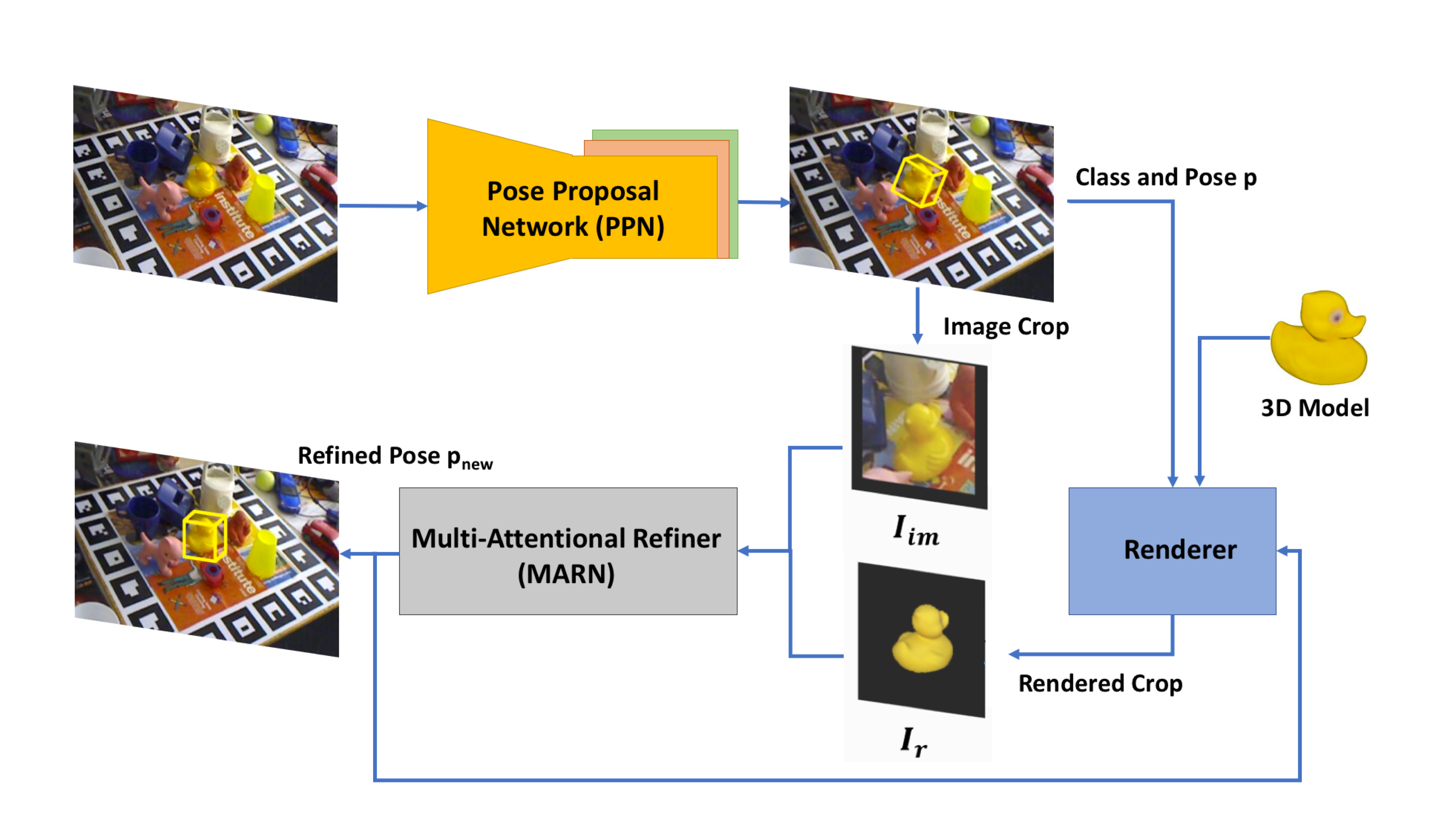}
\caption{An overview of our proposed approach, consisting of a pose proposal module and a pose refinement module. The pose proposal module (PPN), outputs an object classification and an initial pose estimation from RGB inputs. The pose
refinement module consists of a differentiable renderer and an iterative refiner called MARN. The renderer initializes the rendered crop of the detected object using its initial pose estimate and its 3D model. The refinement step utilizes a hybrid representation of the initial render and the input image, combining visual and flow features, and integrates a multi-attentional block to highlight important features, to learn an accurate transformation between the predicted pose and the actual, observed pose.} 
\label{fig:overview}
\end{figure}

This paper presents a novel, end-to-end 6D pose estimation approach from RGB inputs. 
In \figurename~\ref{fig:overview}, we show an overview of our approach.
First, the Pose Proposal Network (PPN) extends the region proposal framework to classify and regress initial estimates of the rotations and translations of objects present in the RGB input. Notably, our proposed PPN method requires no additional steps, unlike methods that use P$n$P \cite{fischler1981random} or matching with pre-engineered codebook. 
Second, the pose refinement module consists of a differentiable renderer and a Multi-Attentional Refinement Network (MARN). 
MARN can be depicted as two main components: first, visual features from both the input crop and the rendered crop are fused using the flow vectors to learn better object representations. Second, a spatial multi-attention block highlights discriminative feature parts, insulating the network from adverse noise and occlusion effects. MARN is designed to allow iterative refinement; MARN outputs an estimated pose that directly maps to MARN's input. In our experiments, we typically found that the greatest performance gains occurred within a couple of refinement iterations. 

Finally, our entire pipeline is trained end-to-end and achieves state-of-the-art results across a range of experiments and datasets. 

In summary, our work makes the following contributions: 

\begin{enumerate}

 \item An end-to-end 6D pose estimation approach that outperforms state-of-the-art RGB-based methods on three commonly used benchmarks.
 \item A pose proposal network (PPN) that is fully-CNN-based, yielding fast and accurate pose estimations in a single pass from RGB images, without additional steps.
 \item A pose refinement network (MARN) that uses a hybrid intermediate representation of the input image and the initial pose estimation by combining visual and flow features to learn an accurate transformation between the predicted object pose and the actual observed pose. 
 \item The integration of a spatial multi-attentional block that highlights important feature parts, making the refinement process more robust to noise and occlusion.
 
\end{enumerate}

\section{Related Work}
\label{section:related-work}
6D pose estimation has a long and storied history \cite{wohler20123d}, but, due to space constraints, we will limit this section to methods that use RGB inputs.
Traditional RGB methods for pose estimation typically match detected local keypoints or hand-crafted features with known object models \cite{aubry2014seeing,collet2011moped,lowe2004distinctive,wagner2008pose}. These methods maintain scale and rotation invariance, and hence are often faster and more robust to occlusion. However, they become unreliable with low-texture objects, low-resolution or noisy inputs \cite{gorbett2020utilizing}. Deep learning methods tend to be more robust to these issues. More recent variants of these methods mostly rely on deep learning to either learn feature representations or create 2D-3D correspondences \cite{brachmann2014learning,krull2015learning}. 

Most existing RGB-based methods \cite{kehl2017ssd,oberweger2018making,peng2019pvnet,rad2017bb8} take advantage of deep learning techniques used for object detection \cite{girshick2015fast,liu2016ssd,redmon_you_2016} or image segmentation \cite{long2015fully} and leverage them for 6D pose estimation.
For example, one technique involves utilizing CNNs to extract object keypoints, and solving the 6D poses using P$n$P \cite{rad2017bb8,tjaden2017real}. Sundermeyer et.al \cite{sundermeyer2018implicit} utilizes an encoder-decoder that learns feature vectors and matches them to a pre-generated codebook of feature vectors to determine the 6D pose of an object. Our work is different from these methods in that we integrate a pose estimator based on a region proposal framework, that estimates object poses in a single forward pass through the network with no additional codebook matching steps. Kehl et.al \cite{kehl2017ssd} and Tekin et.al \cite{tekin2018real} use region proposal framework to detect objects within the input image and then use additional steps (such as P$n$P \cite{fischler1981random}) to solve their poses. Though our approach also integrates a pose estimator inspired from region proposal frameworks, our work leverages the framework in an encoder/multi-decoder network for 6D pose estimation and extends it into a novel end-to-end pose estimation and refinement network. 

6D pose refinement has been utilized to improve the performance of several pose estimation methods \cite{wang2019densefusion,xiang2017posecnn}. Recent refinement methods have been deep-learning based \cite{wang2019densefusion,manhardt2018deep}, relying on CNNs to predict a relative transformation between the initial pose prediction and the target pose. Li et.al \cite{li2018deepim} relies on Flownet's \cite{ilg2017flownet} deep feature representation, extracted from the input image and the rendering of the estimated object pose to learn the pose residuals. Though our refiner was inspired from \cite{li2018deepim}, our approach is fundamentally different as it relies on the synergy between the optical flow vectors and the appearance features to capture the pose transformation from the prediction to the target pose. Further, we employ a multi-attentional block that efficiently highlights discriminative feature parts, improving the robustness of our refiner to noise and occlusion.

\section{Methods}
\label{sec:others}
In this paper, we estimate the 6D poses of a set of known objects present in an RGB image. We propose a novel two step approach: First, a pose proposal module (PPN) (\S~\ref{section:pose_proposal_network}) that regresses initial 6D pose proposals from different regions of objects in an RGB image. Second, a pose refinement module, which includes i) a differentiable renderer, that outputs a render of the detected object using its initial pose estimate and 3D model, and ii) a multi-attentional refinement network (MARN) (\S~\ref{section:multi-attential_refinement_network}), to further refine the initial pose estimates.

In the following, the 6D pose is represented as a homogeneous transformation matrix, $p= [R|t] \in \mathcal{SE}(3)$, composed of a rotation matrix $R \in \mathcal{SO}(3)$ and a translation $t \in \mathbb{R}^3$. $R$ can also be represented by a quaternion $q \in \mathbb{R}^4$.

\subsection{Pose Proposal Network} \label{section:pose_proposal_network}

\begin{figure}
\centering
\includegraphics[height=12cm]{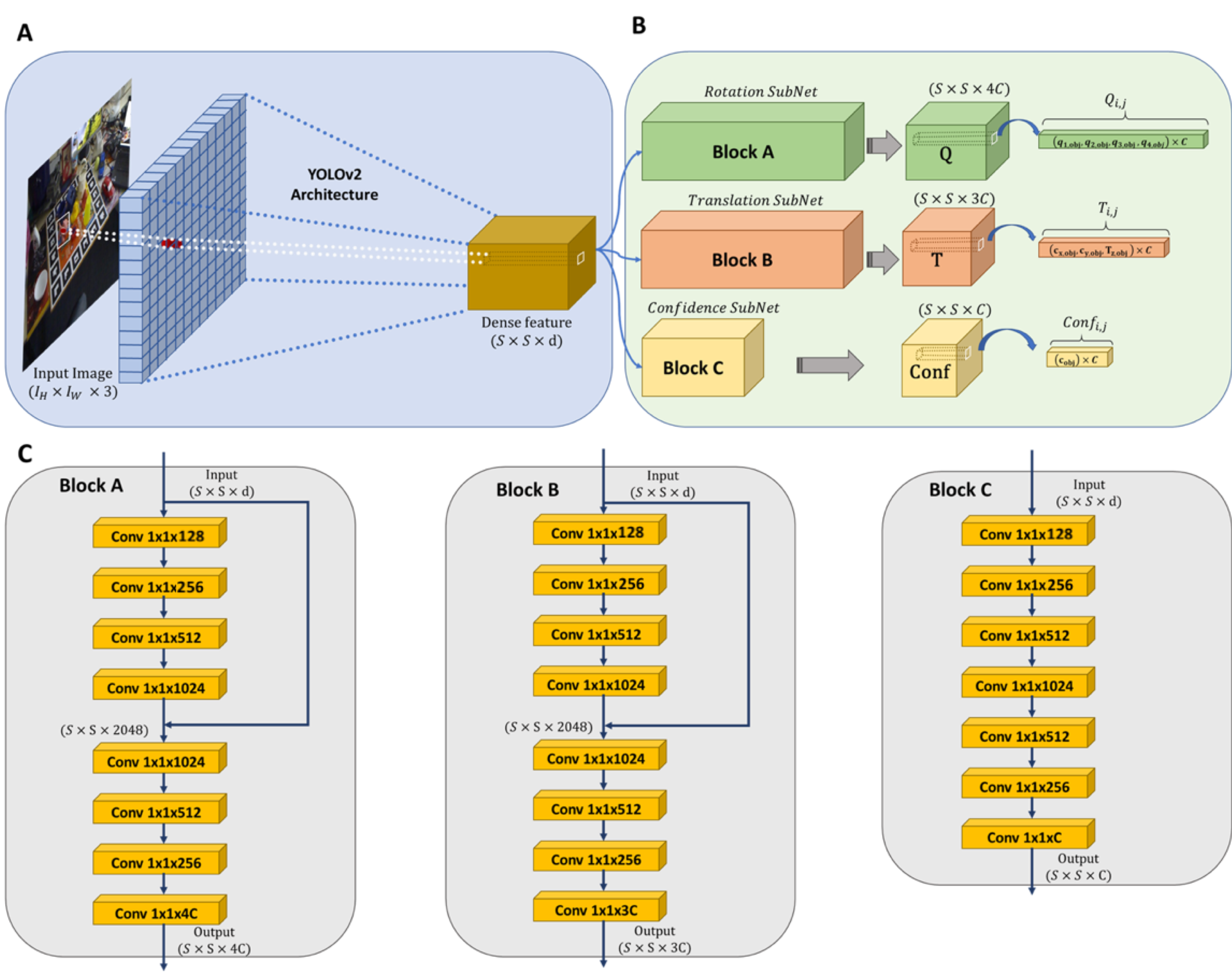}
\caption{Our Pose Proposal Network (PPN) Architecture. The encoder/multi-decoder network takes an RGB image, \textbf{A}. encodes it into high dimensional feature embedding, and \textbf{B}. decodes it into 3 task-specific outputs, which correspond to the rotation, translation, and confidence in the presence of the detected object. \textbf{C}. Architectural details of blocks A, B, and C in our PPN} 
\label{fig:backbone}
\end{figure}

We reframe the object pose estimation as a combined object classification and pose estimation problem, regressing from image pixels to region proposals of object centers and poses. \figurename~\ref{fig:backbone} illustrates our 6D object pose proposal network architecture. Our architecture has two stages: first, a backbone encoder, modeled on the YOLOv2 framework~\cite{redmon2017yolo9000}, extracts high-dimensional region feature representations from the input image. Second, the obtained feature representations are embedded into low-dimensional, task-specific features extracted from three decoders which output three sets of region proposals for translations, rotations, and object centers and classes.We note that, similar to \cite{tekin2018real}, we rely on the YOLOv2 framework (\S~\ref{section:related-work}) to extract feature representations from the input image. However, the application of the YOLOv2 network in our work is fundamentally different in the sense that it serves only the purpose of extracting appearance features from the input image which will be used as input to the second stage consisting of three decoders to ultimately estimate the objects poses.
Specifically, the backbone encoder (\figurename~\ref{fig:backbone}\textbf{A}) produces a dense feature representation $\mathcal{F}$ by dividing the input image into a $S \times S$ grid, each cell of which corresponds to an image block, that produces a set of high dimensional feature embeddings $\{ \mathcal{F}_{i,j} \}$, with $\mathcal{F}_{i,j} \in \mathbb{R}^d$ for each grid cell $(i,j) \in \mathcal{G}^2 \text{ s.t. } \mathcal{G} = \{1, \ldots , S \}$ and $d$ is the embedding size.
$\mathcal{F}$ is decoded by 3 parallel convolutional blocks (as shown in \figurename~\ref{fig:backbone}\textbf{B}) that produce a fixed-size collection of region proposals $\{ (Conf_{i,j}^{o_k},T_{i,j}^{o_k},Q_{i,j}^{o_k})\}$ for each object in the set of target objects $ o_k \in \{o_1, \ldots, o_C\}$, where $C$ is the number of target objects. 
The detailed architectures of the three blocks are depicted in supplemental material.

\textbf{Block A:} This block is a rotation proposal network that regresses a 4-dimensional quaternion vector $Q_{i,j}^{o_k}$ for each image region and object class.

\textbf{Block B:} This block is a translation proposal network that regresses a 3-dimensional translation vector $T_{i,j}^{o_k}$ for each image region and object class. Rather than predicting the full translation vector $T = [t_x, t_y, t_z]^T$, which can be cumbersome for training as discussed in \cite{xiang2017posecnn}, we regress the object center coordinates in the image space $c = (c_x,c_y)^T$ and the depth component $t_z$. The two remaining components of the translation vector are then easily computed with the camera intrinsics and the predicted information:
\begin{align}\label{t_transf}
\begin{split}
    t_x = \frac{(c_x - p_x) t_z}{f_x}, \\
    t_y = \frac{(c_y - p_y) t_z}{f_y} 
\end{split}
\end{align}
where $f_x$ and $f_y$ denote the focal lengths of the camera, and $(p_x, p_y)$ is the principal point offset. To regress the object's center coordinate, we predict offsets for the 2D coordinates with respect to $(g_x, g_y) \in \mathcal{G}^2$, the top-left corner of the associated grid cell. We constrain this offset to lie between 0 and 1. The predicted center point $(c_x, c_y)$ is defined as:
$ c_x = f(x) + g_x$ and $c_y = f(y) + g_y$ where $f(\cdot)$ is a 1-D sigmoid function.

\textbf{Block C:} This block is a confidence proposal network, which should have high confidence in regions where the object is present and low confidence in regions where it is not. 
Specifically, for each image region, Block C predicts a confidence value for each object class corresponding to the presence or absence of that object's \emph{center} in the corresponding region in the input image. 

\textbf{Duplication Removal:} After the inference of object detection and pose estimation, which is done by one pass through our PPN, we apply non-maximal suppression to eliminate duplicated predictions when multiple cells have high confidence scores for the same object. Specifically, the inference step provides class-specific confidence scores, referring to the presence or absence of the class in the corresponding grid cell. Each grid cell produces predictions in one network evaluation, and cells with low confidence predictions are pruned using a confidence threshold. We then apply non-maximal suppression to eliminate duplicated predictions when multiple cells have high confidence scores for the same object and only consider the predictions with the highest confidence score, assuming either the object center lies at the intersection of two cells or the object is large enough to occupy multiple cells. We specifically measure the similarity of the projected bounding boxes of the 3D models given the predicted poses by computing the overlap score using intersection over union (IoU). Given two bounding boxes with high overlap score, we remove the bounding box that has the lower confidence score.
This step is repeated until all of the non-maximal bounding boxes has been removed for every class. Two projections are considered to be overlapping if the IoU score is larger than 0.3.

\subsection{Multi-Attentional Refinement Network} \label{section:multi-attential_refinement_network}

\begin{figure}[t]
\centering
\includegraphics[height=5.5cm]{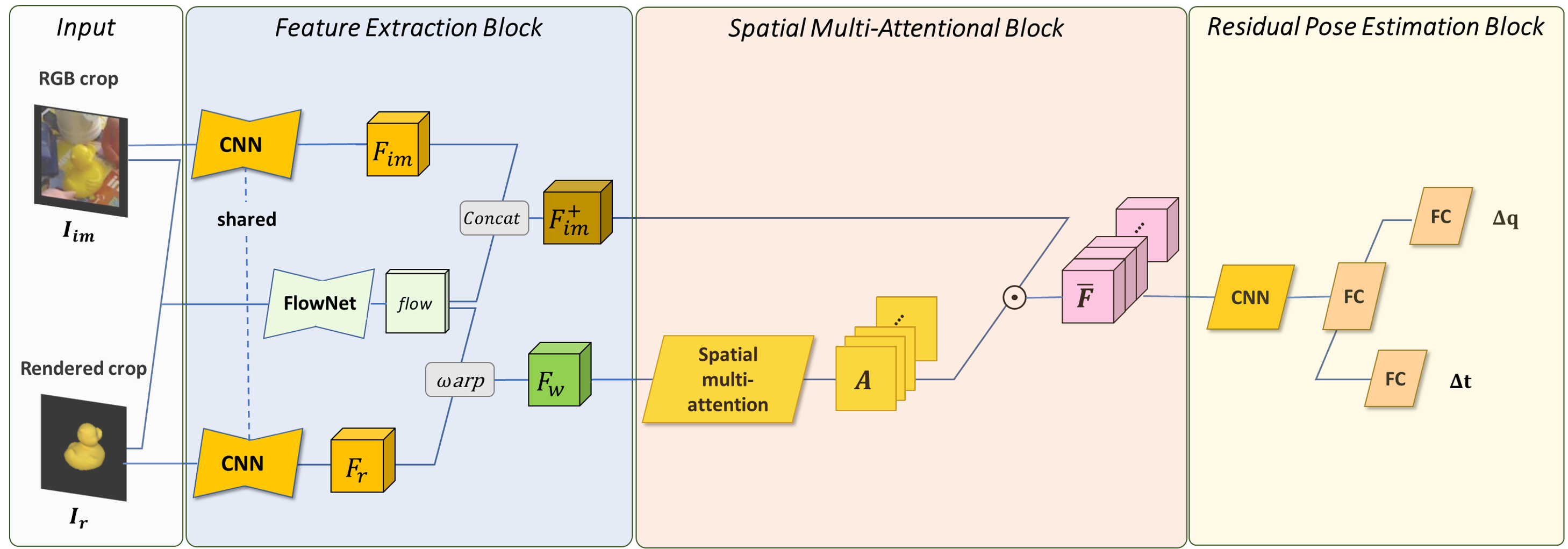}
\caption{Our proposed multi-attentional refinement network (MARN) takes a proposed pose and iteratively refines it. In the context of our pipeline, the initial pose estimate, represented as a render image crop and a real image crop, are input into MARN. First, the network extracts visual feature representations from the inputs and an optical flow estimation between the two inputs (the Feature Extraction Block). Then, multiple attention maps, which correspond to different parts of the target object, are extracted from the flow and render crop features and applied to the feature representation of the real image crop, highlighting the important feature parts (the Spatial Multi-Attentional Block). Subsequently, the highlighted features are used to refine the pose estimate (the Residual Pose Estimation Block). The output refined pose estimate can be input into MARN for iterative refinement}
\label{fig:ref}
\end{figure}

Our proposed multi-attentional refinement network (MARN) iteratively corrects the 6D pose estimation error. Given the success of end-to-end trainable models \cite{ruder2017overview,Chaabane_2020_WACV}, we opt for an end-to-end refinement pipeline. \figurename~\ref{fig:ref} depicts the MARN architecture and illustrates a typical refinement scenario. Two color crops ($I_{im}$ and $I_r$), corresponding to an observed image and an initial pose estimate of the object in the image, are input into MARN, which outputs a pose residual estimate to update the initial predicted pose. This procedure can be applied iteratively, potentially generating finer pose estimation at each iteration. 

\textbf{Input Crops:}\label{input_sampling} 

Input Crops are sampled from a given predicted 6D pose $p$. Crops circumvent the difficulty of extracting visual features from small objects. Two crops, a rendered and an RGB, are generated. Images are cropped under the assumption that only minor refinements are needed. Both crops will be used as input to the refinement network.
The rendered crop is generated by rendering the 3D object model viewed according to the predicted pose $p$. The RGB crop is generated from the original input image. 
We compute a bounding box, that bounds the object's 3D model, projected on the image space using the predicted pose $p$. We pad the bounding box by $epsilon$ pixels for each side to take into account the error introduced by the pose prediction. The enlarged bounding box is then used as a mask applied to the RGB image. Note that the mask cancels out the background, it does not crop the images. 
The images are cropped with a fixed size window $H\times W$, where the crop center corresponds to the object center, as defined by the 2D projection of the predicted pose $p$.
Predicting $(\Delta c_x, \Delta c_y)$ consists of estimating how far the object center is from the image center.

\textbf{Feature Extraction Block:}

MARN refines the estimated pose by predicting the relative transformation to match the rendered view of the object to the observed view in the original image. To this end, MARN's feature extraction block is composed of two different networks: 1) a visual feature embedding network that captures visual features of the object, and 2) a flow estimation network that estimates the object ``motion" between the rendered image and the observed image. 
The network takes two input crops: $I_r \in \mathbb{R}^{H \times W \times 3}$ and $I_{im} \in \mathbb{R}^{H \times W \times 3}$. Both crops are processed through the shared visual feature embedding network to extract visual feature representations $F_{im} \in \mathbb{R}^{H \times W \times d_{em}}$ for the image crop and $F_r \in \mathbb{R}^{H \times W \times d_{em}}$ for the render crop. Each pixel location of the embedding is a $d_{em}$-dimensional vector that represents the appearance information of the input image at the corresponding location. Simultaneously, the flow estimation network, based on the FlowNetSimple architecture \cite{dosovitskiy2015flownet}, produces the optical flow between the rendered image and the observed image. 

Subsequently, the visual feature map $F_r$, extracted from the render crop, is warped toward the visual feature map of the image crop $F_{im}$, guided by the flow information. 
Specifically, the warping function $\mathcal{W}$, extracted from the Flow estimation network, computes a new warped feature map $F_w$ from the input $F_r$ following the flow vectors $flow_{r \xrightarrow{}im} \in \mathbb{R}^{H \times W \times 2}$: 

\begin{align}
  F_w = \mathcal{W}(F_r, flow_{r \xrightarrow{}im})
\end{align}
Following \cite{ilg2017flownet}, the warping operation is a bilinear function applied on all locations for each channel in the feature map. The warping in one channel $l$ is performed as:

\begin{align}
  F^l_w(\boldsymbol{x_w}) = \sum_{\boldsymbol{x_r}} \mathcal{I}(\boldsymbol{x_r},\boldsymbol{x_w}+\delta \boldsymbol{x_w}) F^l_r(\boldsymbol{x_r})
\end{align}
where $\mathcal{I}$ is the bilinear interpolation kernel, $\boldsymbol{x_r} = (x_r, y_r)^T$ is the 2D coordinates in the visual feature embedding $F_r$, and $\boldsymbol{x_w} = (x_w, y_w)^T$ is the 2D coordinates in the visual feature embedding $F_w$. For backpropagation, gradients to the input CNN and flow features are computed as in \cite{ilg2017flownet}.
Furthermore, the estimated optical flow $flow_{r \xrightarrow{}im}$ is concatenated with the feature map extracted from the image crop $F_{im}$ to produce $F^+_{im} \in \mathbb{R}^{H \times W \times (d_{em} +2)}$.

\textbf{Spatial Multi-Attention Block:}

Estimating an object’s relative transformation between two images ﬁrst requires successful localization of the target object within the two inputs. MARN handles this in the spatial multi-attention block by localizing discriminative parts of the target object with spatial multi-attention maps, which robustly localize discriminative parts of the target. Therefore when the target is partially occluded, our multiple attention module can adaptively detect the visible parts while ignoring the occluded parts.
Attention maps $\mathcal{A} = \{a_1, a_2, \ldots, a_N\}$, where $a_i \in \mathbb{R}^{H \times W}$ for $i \in \{1, \ldots, N \}$ and $N$ is the number of attention maps, are extracted by generating summarized feature maps $s_i \in \mathbb{R}^{H \times W}$ for $i \in \{1, \ldots, N \}$ by applying two $1\times1$ convolutional operations to feature map $F_w$, extracted by the feature extraction block. 
Each attention map $a_i \in \mathcal{A}$, corresponding to a discriminative object part, is obtained by normalizing the summarized feature map $s_i$ using softmax:
\begin{align}
a_i = \frac{ \exp{(s_i)}}{\sum^{H}_{h=1} \sum^{W}_{w=1} \exp{(s_{i,h,w})}}, \quad i=1 , \ldots, N
\end{align}

Finally, the attention map $a_i$ and the feature map $F^+_{im}$ are element-wisely multiplied to extract the attentional feature map $\bar{F}_i$:

\begin{align}
\bar{F}_i = A_i \cdot F^+_{im}, \quad i=1 , \ldots, N
\end{align}
where $A_i \in \mathbb{R}^{H \times W \times (d_{em}+2)}$ is the replication of the attention map $a_i$, $(d_{em}+2)$ times to match the dimensions of $F^+_{im}$. $\bar{F} \in \mathbb{R}^{H \times W \times (d_{em}+2)N}$ is the final extracted multi-attentional feature representation obtained by concatenating the attentional feature maps $\{\bar{F}_i\}_{i=1,\ldots,N}$.
Inspired by \cite{prakash2019repr}, we add a regularization term to the total loss function to discourage multiple attention maps locating the same discriminative object part. The regularization emphasizes orthogonality among the attention maps: 
\begin{align}
\mathcal{L}_{orth} = \norm{\tilde{A}^T\tilde{A} - I}_2
\end{align}
where $\tilde{A} = [\tilde{a}_1, \ldots, \tilde{a}_N] \in \mathbb{R}^{HW \times N} $ and $\tilde{a}_i \in \mathbb{R}^{HW}$ is the vectorized attention map of $a_i$. 


\textbf{Residual Pose Estimation Block:}

This block processes the residual pose estimation. First, the embedding space of the extracted feature map $\bar{F}$ is reduced the from $(d_{em}+2)N$ to $8$ with three $3 \times 3$ convolutional operations. The resulting feature map is then fed into one fully connected layer, whose output is then fed into two separate fully connected and final output layers, one corresponding to the regressed rotation and the other corresponding to the translation. As explained in \S \ref{input_sampling}, MARN outputs an estimated relative rotation quaternion $\Delta q \in \mathbb{R}^4$ and a relative translation $[\Delta c_x, \Delta c_y, \Delta t_z]^T$. The refined pose prediction is then computed with regard to the the initial pose prediction $\hat{p} = [\hat{R}|\hat{t}]$ using $c_{x, new} = c_x + \Delta c_x$, $c_{y, new} = c_y + \Delta c_y$, $\hat{t}_{z,new} = \hat{t}_z + \Delta t_z$, and $\hat{R}_{new} = \Delta R \ast \hat{R}$, where $(c_x,c_y)$ is the center of the object in the image space using $\hat{p}$, $\ast$ is the matrix multiplication and $\Delta R$ is the relative rotation matrix obtained from $\Delta q$. $\hat{t}_{x,new}$ and $\hat{t}_{y,new}$ are then computed using \eqref{t_transf}.

\subsection{Losses:} \label{section:losses}
In order to achieve accurate pose estimation, we must provide a criterion which quantifies the quality of the predicted pose. 
The different components of our approach are trained jointly in an end-to-end fashion with a multi-task learning objective:
 
 \begin{align}\label{total_loss}
 \begin{split}
 \mathcal{L}_{total} &= \mathcal{L}_{PPN} + \mathcal{L}_{MARN}\\
 &= \alpha \mathcal{L}_{pose} + \beta \mathcal{L}_{conf} + \gamma \mathcal{L}_{ref}+ \kappa \mathcal{L}_{orth}
 \end{split}
\end{align}

where $\alpha$, $\beta$, $\gamma$ and $\kappa$ are weight factors.
Our multi-task learning objective is composed of four loss functions. First, a composite $\mathcal{L}_2$ loss function to optimize the PPN pose and center detection parameters:

\begin{align}\label{PPN_loss}
 \begin{split}
 \mathcal{L}_{PPN} &= \alpha \mathcal{L}_{pose} + \beta \mathcal{L}_{conf}\\
 \textrm{where} \quad \mathcal{L}_{pose} &= \underset{x \in \mathcal{M}_s}{avg} \norm{(Rx+t)-(\hat{R}x+\hat{t})}_2\\
 \textrm{and} \quad\mathcal{L}_{conf} &= \norm{conf_{gt} - conf_{pr}}_2
 \end{split}
\end{align}
where $\norm{\cdot}_2 $ is the $L_2$ norm. $\mathcal{L}_{conf}$ is the loss term used to train the confidence block. $\mathcal{L}_{pose}$ is the loss term used to train the pose regression. $\mathcal{L}_{pose}$ is similar to the average distance (ADD) measure (further discussed in \S~\ref{experiments}). $p = [R|t]$ is the ground truth pose and $\hat{p} = [\hat{R}|\hat{t}]$ is the estimated pose. $\hat{R}$ and $R$ are the rotation matrices computed from the predicted quaternion $\hat{q}$ and the ground truth quaternion $q$, respectively. $conf_{gt}$ and $conf_{pr}$ are the ground-truth and the predicted confidence matrix, respectively. $\mathcal{M}_s \in \mathbb{R}^{M\times3}$ is a set of points sampled from the CAD model. 
$\mathcal{L}_{pose}$ is only used for asymmetric objects. To handle symmetric objects, we instead use:
\begin{align}
 \mathcal{L}_{pose,sym} &= \underset{x_1 \in \mathcal{M}}{avg} \min_{x_2 \in M} \norm{(Rx_1+t)-(\hat{R}x_2+\hat{t})}_2
\end{align}

Second, MARN's loss function is defined as:
\begin{align}
  \begin{split}
  \mathcal{L}_{MARN} &= \gamma \mathcal{L}_{ref} + \kappa \mathcal{L}_{orth}\\
  \textrm{where} \quad \mathcal{L}_{ref} &= \underset{x \in \mathcal{M}_s}{avg} \norm{(Rx+t)-(\hat{R}_{new}x+\hat{t}_{new})}_2
  \end{split}
\end{align}

$\mathcal{L}_{ref}$ is the same loss term used in PPN. Symmetric objects are handled similarly to PPN. $\hat{R}_{new}$ and $\hat{t}_{new}$ are the refined rotation and translation estimates.

$\mathcal{L}_{orth}$ is a regularization term used to discourage multiple attention maps locating the same discriminative object part. The regularization emphasizes orthogonality among the attention maps as proposed by \cite{prakash2019repr}:

\begin{align}
\mathcal{L}_{orth} = \norm{\tilde{A}^T\tilde{A} - I}_2
\end{align}
where $\tilde{A} = [\tilde{a}_1, \ldots, \tilde{a}_N] \in \mathbb{R}^{HW \times N} $ and $\tilde{a}_i \in \mathbb{R}^{HW}$ is the vectorized attention map of $a_i$.
\subsection{Architectural and Training Details:} \label{section:training_details}
Below we present details about both our training procedures and system architecture. These details specifically pertain to experiments which follow. 

Our model is optimized with Adam optimizer with weight factors ($\alpha$, $\beta$ , $\gamma$, $\kappa$) set to ($0.1$, $0.05$, $0.1$, $0.01$). 

\subsubsection{PPN:}

The backbone encoder in PPN consists of 23 convolution layers and 5 max-pooling layers, following the YOLOv2 architecture \cite{redmon2017yolo9000}. Additionally, we add a pass-through layer to transfer fine-grained features to higher layers. Our model is initialized with pre-trained weights from YOLOv2, with the remaining weights being randomly initialized. Input images are resized to $416\times416$ and split into $13\times 13$ grids ($S=13$). The feature embedding size of the backbone network, $d$, is set to be equal to $1024$.

Initially, we use an additional weight factor, $\lambda $, that we apply to the confidence block output. Specifically, PPN is trained with $\lambda$ set to $5$ for the cells that contain target objects and $0.5$ otherwise. This circumvents convergence issues with the confidence values because otherwise the early stages of training tend to converge on all zeros (since the number of cells that contain objects is likely to be much smaller than the cells that do not). In later training stages, $\lambda$ is updated to penalize false negatives and false positives equally ($\lambda = 1$ for all cells). The number of points $M$, in the set of 3D model points $\mathcal{M}_s$, is set to $10,000$ points.

\subsubsection{MARN:}

For our visual feature embedding network, we use a Resnet18 encoder pre-trained on ImageNet followed by 4 up-sampling layers as the decoder. During training, the two networks are fine-tuned with shared weight parameters. We set the embedding size of the extracted features from the visual feature embedding network, $d_{em}$, to be equal to $32$.
The flow estimation network is the FlowNetS architecture populated with pre-trained weights following \cite{dosovitskiy2015flownet}. The network weights are frozen for the first two training epochs and unfrozen in later epochs. Once the weights are unfrozen, the component is trained in an end-to-end manner along with the other MARN components. The initial weight freeze increases training stability and ensures the output of the flow estimation network is meaningful. FlowNet output is up-sampled to match the input image crops. After a hyperparameter search, the padding offset for the mask $\epsilon$ was set to 10 pixels and the cropping window size is set to $H \times W = 256 \times 256$ applied to the original input image. 
Pose perturbations are used to create training data by adding angular perturbations ($5\deg$ to $45\deg$) and/or translational perturbations ($0$ to $1$ relative to the object's diameter) to obtain a new noisy pose and rendering an image. The network is then trained to estimate the target output which is the relative transformation between the perturbed pose and the ground-truth pose.

\section{Experiments:}\label{experiments}

The full model was implemented with PyTorch and all experiments were conducted on a Ubuntu server with a TITAN X GPU with $12$ GB of memory. All models and code will be made publicly available upon publication.

In this section, our pose estimation models are compared against state-of-the-art RGB-based methods across three datasets, YCB-Video (\S~\ref{section:ycb}), LINEMOD (\S~\ref{section:linemod}), and LINEMOD Occlusion (\S~\ref{section:linemod-occlusion}), and obtain state-of-the-art results on all datasets, with competitive runtimes. Given a $480 \times 640$ input image, PPN alone runs at 50 fps and the full model runs at 10 fps, with two refinement iterations, which is efficient for real-time pose estimation.  We also show in Supplemental material that our PPN alone has competitive performance when compared with methods that do use such information.

\subsection{Evaluation Metrics:}
Two standard performance metrics are used. First, the 2D-projection error, analogously to \cite{peng2019pvnet}, measures the average distance between the 2D projections in the image space of the 3D model points, transformed using the ground-truth pose and the predicted pose. The pose estimate is considered to be correct if it is within a selected threshold. 2D-Proj denotes the percentage of correctly estimated poses using a 2D Projection Error threshold set to 5 pixels. For symmetric objects, the 2D projection error is computed against all possible ground truth poses, and the lowest value is used. The second metric, Average 3D distance (ADD) \cite{hinterstoisser2012model}, measures the average distance between the 3D model points transformed using the ground-truth pose and the predicted pose. For symmetric objects, we use the closet point distance, referred to as ADD-S in \cite{xiang2017posecnn}. In our experiments, we denote as ADD(-S), following \cite{xiang2017posecnn}, the metric that measures the percentage of correctly estimated poses using a ADD(-S) threshold. Unless specified, in our experiments the threshold is set to 10\% of the 3D model diameter. When evaluating on the YCB-Video dataset, we also report the ADD(-S) AUC as proposed in~\cite{xiang2017posecnn}.

\subsection{Evaluation on YCB-Video Dataset:} \label{section:ycb}
 
The YCB-Video dataset \cite{xiang2017posecnn} has 21 objects \cite{calli2015ycb} across 92 video sequences. In our experiments, we divide the data as in \cite{xiang2017posecnn}, using 80 sequences for training and 20 sequences for testing. We augment our training with 80k synthetically rendered images released by \cite{xiang2017posecnn}. Pose predictions on the test set was refined with four MARN iterations.

\setlength{\tabcolsep}{4pt}

\begin{table}[t]
\scriptsize
\begin{center}
\caption{Comparison of our approach with state-of-the-art RGB-based methods on \textbf{YCB-Video dataset} in terms of 2D-Proj, ADD AUC and ADD(-S) metrics, averaged over all object classes for each method. We use a threshold of 2 cm for the ADD(-S) metric}
\label{table:comp_YCB}
\begin{threeparttable}
\begin{tabular}{|c|cccc|}
\hline 
Methods & HMap\cite{oberweger2018making} & PVNet\cite{peng2019pvnet} & DeepIM$^{\dagger}$\cite{li2018deepim} & OURS$^{\dagger}$\\
\hline
2D-Proj&39.4&47.4&-&\textbf{55.6}\\
\hline
ADD AUC &72.8&73.4&81.9&\textbf{83.1}\\
\hline
ADD(-S) ($<2cm$)&-&-&71.5&\textbf{73.6}\\

\hline
\end{tabular}
 \begin{tablenotes}[para,flushleft]
 $^{\dagger}$ denotes methods that deploy refinement steps.
 \end{tablenotes}
 \end{threeparttable}
\end{center}
\end{table}

\setlength{\tabcolsep}{1.4pt}

\subsubsection{Results:}

The results in \tablename~\ref{table:comp_YCB} suggest that our approach significantly outperforms state-of-the-art RGB-based methods with an average 2D-Proj accuracy of 55.6\%. Compared to DeepIM \cite{li2018deepim}, which also deploys refinement steps, our proposed approach achieves better performance by a margin of 1.2\% and 2.1\% in terms of ADD AUC and ADD(-S) respectively. Detailed results, broken down by object, can be found in the supplemental material. Our approach achieves the best results in 12 object classes out of 21 compared to other methods.

\textbf{Detailed Results on the YCB-Video Dataset:}
In \tablename~\ref{table:comp_YCB_det}, we show detailed pose estimation results on the YCB-Video dataset\cite{xiang2017posecnn} in terms of ADD AUC. Our  approach achieves the best results in 12 object classes out of 21 compared to other methods. DeepIM, surpasses other methods on 6 object classes out of 21, and HMap outperforms other methods on 4 object classes.

\setlength{\tabcolsep}{4pt}
\begin{table}
\scriptsize
\begin{center}
\caption{Detailed results of our approach and other existing RGB-based methods on the different objects of the YCB-Video dataset in terms of ADD AUC}
\label{table:comp_YCB_det}
\begin{threeparttable}
\begin{tabular}{|l|c|c|c|c|}
\hline 
Methods & HMap\cite{oberweger2018making} & PVNet\cite{peng2019pvnet} &DeepIM$^{\dagger}$\cite{li2018deepim} & OURS$^{\dagger}$\\
\hline

002-master-chef-can &\textbf{81.6}&-&71.2&72.1\\
003-cracker-box&\textbf{83.6}&-&\textbf{83.6}&81.7\\
004-sugar-box&82.0&-&94.1&\textbf{95.7}\\
005-tomato-soup-can &79.7&-&86.1&\textbf{88.2}\\
006-mustard-bottle&91.4&-&91.5&\textbf{94.8}\\
007-tuna-fish-can &49.2&-&87.7&\textbf{88.2}\\
008-pudding-box &\textbf{90.1}&-&82.7&80.2\\
009-gelatin-box&93.6&-&91.9&\textbf{94.5}\\
010-potted-meat-can &79.0&-&76.2&\textbf{82.6}\\
011-banana&51.9&-&\textbf{81.2}&78.7\\
019-pitcher-base &69.4&-&\textbf{90.1}&87.7\\
021-bleach-cleanser &76.1&-&\textbf{81.2}&78.1\\
024-bowl*&76.9&-&81.4&\textbf{83.4}\\
025-mug&53.7&-&81.4&\textbf{81.7}\\
035-power-drill&82.7&-&85.5&\textbf{87.8}\\
036-wood-block*&55.0&-&81.9&\textbf{83.7}\\
037-scissors&65.9&-&60.9&\textbf{67.4}\\
040-large-marker&56.4&-&\textbf{75.6}&71.1\\
051-large-clamp*&67.5&-&74.3&\textbf{75.2}\\
052-extra-large-clamp*  &53.9&-&\textbf{73.3}&71.3\\  
061-foam-brick*&\textbf{89.0}&-&81.9&82.2\\
\hline
MEAN&72.8&73.4&81.9&\textbf{83.1}\\
\hline
\end{tabular}
 \begin{tablenotes}[para,flushleft]
 $^{\dagger}$ denotes methods that deploy refinement steps.\\
 * denotes symmetric objects.
  \end{tablenotes}
  \end{threeparttable}
\end{center}
\end{table}
\setlength{\tabcolsep}{1.4pt}

\textbf{Ablation Study of The Refiner on YCB-Video Dataset:}

We performed an ablation study on MARN's components (detailed in \S~\ref{section:multi-attential_refinement_network}) to measure the effect of each of its components. In all, we test four variants: In variant 1, MARN only uses visual features extracted from the two input crops. In variant 2, MARN uses the flow estimation features but not the attention component, instead fusing the extracted feature map $F^+_{im}$ and the warped feature map $F_w$ with simple concatenation. In variant 3, spatial attention is added, but only a single attention map is used. Variant 4 is the production variant of MARN. Each variant refined the pose 4 times. 
We break down the results of the ablation study in \tablename~\ref{table:ablation}. First, we notice that variant 1 refinement, though the simplest, still improves the pipeline performance significantly by a margin ADD(-S) of 5.2\%. This finding proves that visual features help in capturing the relative transformation between two inputs, and thus helps refine the pose. Variant 2, which adds in optical flow estimation improves the performance of our refiner by 2.3\% over variant 1. We conjecture that the predicted flow ensures that the network learns to exploit the relationship between both crops and thus capture the relative transformation of the object between them. Variants 3 and 4 show that the addition of attention maps helps to improve the performance of the refiner. The improvement of variant 4 over variant 3 demonstrates that multiple attention maps help achieve better performance than a single attention map. We suspect the ability of multiple attention maps to capture various salient parts of the objects helps the model highlight important features, and makes the refinement process robust to various degrees of occlusion in the dataset. 

\setlength{\tabcolsep}{4pt}
\begin{table}[t]
\scriptsize
\begin{center}
\caption{Results of the ablation study on different components of MARN on \textbf{YCB-Video dataset}. We use the same 2cm threshold for ADD(-S). AUC means ADD(-S) AUC. Each variant was refined with 4 iterations}
\label{table:ablation}

\begin{tabular}{|c|c|c|c|c|c|}
\hline 
Experiments& flow vectors & visual features & Attention maps& ADD(-S)&AUC \\
\hline
Variant 1 &None&\checkmark&None&63.7&77.2 \\
\hline
Variant 2 &\checkmark&\checkmark&None&68.9&79.8 \\
\hline
Variant 3 &\checkmark&\checkmark&single&71.2&81.9 \\
\hline
Variant 4&\checkmark&\checkmark&multiple&\textbf{73.6}&\textbf{83.1}\\
\hline
\end{tabular}
\end{center}
\end{table}
\setlength{\tabcolsep}{1.4pt}

\subsection{Evaluation on LINEMOD Dataset:} \label{LINEMOD} \label{section:linemod}

LINEMOD \cite{hinterstoisser2012model} contains 15,783 images of 13 objects, and includes 3D models of the different objects. Each image is associated with a ground truth pose for a single object of interest. The objects of interest are considered as texture-less objects, which makes the task of pose estimation challenging.
The train/test split is chosen following \cite{brachmann2016uncertainty} --- ~$200$ images per object are used in the training set and $1,000$ images per object in the testing set.
When using the LINEMOD dataset, we opt for online data augmentation during training, to avoid over-fitting. Using this method, random in-plane translations and rotations are applied to the image along with random hues, saturations, and exposures. Finally, we change the images by replacing the background with random images from the PASCAL VOC dataset \cite{everingham2010pascal}. Note that for testing on the LINEMOD dataset, two MARN iterations were used for refinement. 

\setlength{\tabcolsep}{4pt}
\begin{table}[b]
\scriptsize
\begin{center}
\caption{Results of our approach compared with state-of-the-art RGB-based methods on the \textbf{LINEMOD dataset} in terms of ADD(-S) and 2D-Proj metrics. We report percentages of correctly estimated poses averaged over all object classes}
\label{table:comp_LINEMOD}
\begin{threeparttable}
\begin{tabular}{|c|c|c|c|c|c|c|}
\hline 
Method & Tekin\cite{tekin2018real}&PVNet\cite{peng2019pvnet} &SSD6D$^{\dagger}$\cite{kehl2017ssd}&DeepIM$^{\dagger}$\cite{li2018deepim}&OURS$^{\dagger}$ \\
\hline
ADD(-S) &55.95&86.27&79&88.6&\textbf{93.87}\\
2D-Proj&90.37&99.0&-&97.5&\textbf{99.19}\\
\hline
\end{tabular}
 \begin{tablenotes}[para,flushleft]
 $^{\dagger}$ denotes methods that deploy refinement steps.
 \end{tablenotes}
 \end{threeparttable}
\end{center}
\end{table}
\setlength{\tabcolsep}{1.4pt}
\subsubsection{Results:}

As shown in \tablename~\ref{table:comp_LINEMOD}, our approach achieves better results than other RGB-based methods in terms of ADD(-S), with an average accuracy of 93.87\% accuracy compared to an average accuracy of 88.6\% for DeepIM, the second best performing method. Detailed results are shown in the supplemental material. Compared with other methods, our approach had the highest performance on 9 of the 13 object classes. Some examples of pose estimation results using the proposed approach on the LINEMOD dataset are shown in \figurename~\ref{fig:rslts_qualt}. 

\textbf{Detailed Results on the LINIEMOD Dataset:}
\subsection{Detailed Results on the LINIEMOD Dataset}
In \tablename~\ref{table:comp_linemod}, we compare our approach with existing state-of-the-art methods: Tekin\cite{tekin2018real}, PVNet\cite{peng2019pvnet}, BB8\cite{rad2017bb8}, SS6D\cite{kehl2017ssd} and DeepIM\cite{li2018deepim} on LINEMOD dataset\cite{hinterstoisser2012model}. Compared  with  other  methods,  our  approach  had the  highest  performance  on  9  of  the  13  object  classes, PVNet had the best performance on 2 object classes, and SSD6D had the best performance on 2 object classes.
\setlength{\tabcolsep}{4pt}
\begin{table}
\scriptsize
\begin{center}
\caption{Detailed Results of our approach and other existing RGB-based methods on the different objects of the LINEMOD dataset in terms of ADD metric}
\label{table:comp_linemod}
\begin{threeparttable}
\begin{tabular}{|l|c|c|c|c|c|c|}
\hline 

Method & Tekin\cite{tekin2018real}&PVNet\cite{peng2019pvnet} &BB8$^{\dagger}$\cite{rad2017bb8}&SSD6D$^{\dagger}$\cite{kehl2017ssd}&DeepIM$^{\dagger}$\cite{li2018deepim} & OURS$^{\dagger}$ \\
\hline
ape &21.62 &43.62&40.4&65&77&\textbf{84.47} \\
benchvise  &81.80&\textbf{99.90}&91.8&80&97.5&98.71\\
cam &36.57&86.86&55.7&78&93.5&\textbf{93.73}\\
can &68.80&95.47&64.1&86&96.5&\textbf{97.84}\\
cat &41.82&79.34&62.6&70&82.1&\textbf{87.33}\\
driller &63.51&96.43&74.4&73&95&\textbf{96.91}\\
duck &27.23&52.58&44.30&66&77.7&\textbf{88.45}\\
eggbox* &69.58&99.15&57.8&\textbf{100}&97.1&98.49\\
glue* &80.02&95.66&41.2&\textbf{100}&99.4&99.5\\
holepuncher &42.63&81.92&67.20&49&52.8&\textbf{84.53}\\
iron &74.97&98.88&84.7&78&98.3&\textbf{99.10}\\
lamp &71.11&\textbf{99.33}&76.5&73&97.5&98.74\\
phone &47.74&92.41&54.0&79&87.7&\textbf{92.53}\\
\hline
MEAN &55.95&86.27&62.7&79&88.6&\textbf{93.87}\\

\hline
\end{tabular}
 \begin{tablenotes}[para,flushleft]
 $^{\dagger}$ denotes methods that deploy refinement steps.\\
 * denotes symmetric objects.
  \end{tablenotes}
  \end{threeparttable}
\end{center}
\end{table}
\setlength{\tabcolsep}{1.4pt}

\textbf{Ablation Study of the refiner on the LINEMOD Dataset}
In \tablename~\ref{table:ablation}, we report the results of ablation study on LINEMOD dataset. The ablation study is similar to the one conducted in the main paper on YCB-Video dataset. The results in \tablename~\ref{table:ablation} suggest that each component iteratively improves the refinement results, highlighting their effectiveness, but the full importance of each method may be somewhat muted, compared to the results on the YCB-Video dataset, since the experiment took place on the LINEMOD dataset, where accuracy is near the dataset ceiling. 

\setlength{\tabcolsep}{4pt}
\begin{table}
\scriptsize
\begin{center}
\caption{Results of the ablation study on different components of our refinement network MARN on LINEMOD dataset}
\label{table:ablation}
\begin{tabular}{|c|c|c|c|c|c|}
\hline 
Experiments& flow features & CNN features & Attention maps& ADD&2D-Reproj \\
\hline
Variant 1 &None&\checkmark&None&87.32&96.59 \\
\hline
Variant 2 &\checkmark&\checkmark&None&89.17&97.99 \\
\hline
Variant 3 &\checkmark&\checkmark&single&91.28&98.56 \\
\hline
Variant 4&\checkmark&\checkmark&multiple&\textbf{93.87}&\textbf{99.19}\\

\hline
\end{tabular}
\end{center}
\end{table}
\setlength{\tabcolsep}{1.4pt}

\subsection{Evaluation on Occlusion Dataset:} \label{section:linemod-occlusion}

The Occlusion dataset \cite{brachmann2014learning} is an extension of the LINEMOD dataset. Unlike LINEMOD, the dataset is multi-object --- $8$ different objects are annotated in each single image, with objects occluded by each other. Our models are trained with the same online data augmentation procedure described in the LINEMOD dataset (\S~\ref{LINEMOD}), further augmented by adding in image objects extracted from the LINEMOD dataset. Four MARN iterations were used for refinement on the Occlusion dataset. 
The Occlusion Dataset is particularly important because it tests the robustness of our pipeline to occlusion, something the spatial multi-attentional block of MARN was explicitly designed to be robust to. 

\setlength{\tabcolsep}{4pt}
\begin{table}
\scriptsize
\begin{center}
\caption{Comparison of our approach with state-of-the-art RGB-based algorithms on Occlusion in terms of ADD(-S) and 2D-Proj metrics. We report percentages of correctly estimated poses averaged over all object classes}
\label{table:Occlusion}
\begin{threeparttable}
\begin{tabular}{|c|c|c|c|c|c|}

\hline 
Method &HMap\cite{oberweger2018making}&PVNet\cite{peng2019pvnet}&BB8$^{\dagger}$\cite{rad2017bb8} &DeepIM$^{\dagger}$\cite{li2018deepim}&OURS$^{\dagger}$\\
\hline 
ADD(-S) &30.4&40.77&33.88&55.5&\textbf{58.37}\\
\hline
2D-Proj&60.9&61.06&-&56.6&\textbf{65.46}\\
\hline
\end{tabular}
 \begin{tablenotes}[para,flushleft]
 $^{\dagger}$ denotes methods that deploy refinement steps.
  \end{tablenotes}
  \end{threeparttable}
\end{center}
\end{table}

\setlength{\tabcolsep}{1.4pt}

\begin{figure}[t]
\centering
\includegraphics[height=3.7cm]{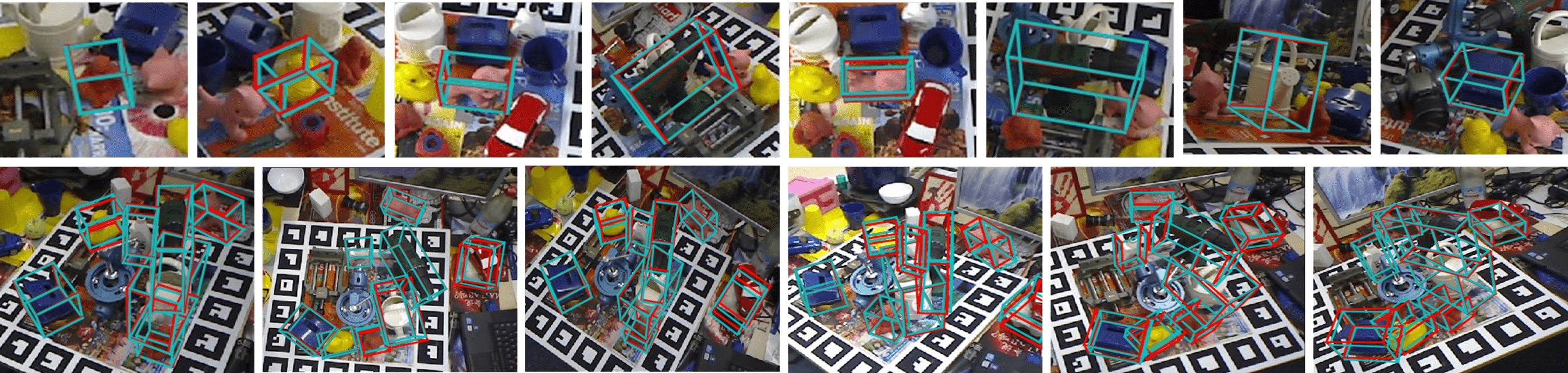}
\caption{Results of poses predicted using the proposed approach. The first row shows results from the LINEMOD dataset. The second row shows results from the LINEMOD Occlusion dataset. In both rows, the cyan bounding boxes correspond to predicted poses and red bounding boxes correspond to ground-truth poses}
\label{fig:rslts_qualt}
\end{figure}

\textbf{Results:}
Results in \tablename~\ref{table:Occlusion} show that, our approach achieves significant improvements over all state-of-the-art RGB-based methods. Specifically, our approach surpasses DeepIM by an ADD(-S) margin of 2.87\% and PVNet by 17.6\%. Furthermore, our approach significantly outperforms HMap, which was explicitly designed to handle occlusion, by an ADD(-S) margin of 27.97\%. The significant improvement in performance on the Occlusion dataset, shows the importance of the different components of our MARN, and mainly the spatial multi-attentional block, in robustly recovering the poses of objects under severe occlusion.
In \figurename~\ref{fig:rslts_qualt}, we show examples of pose estimation results using the proposed approach on Occlusion dataset. Even when most objects are heavily occluded, our approach robustly recovers their poses.

\subsection{PPN Only: An Efficient Pose Estimator for Real Time Applications} \label{section:pose_estimation_from_RGB_alone}

We evaluate the performance of PPN, our pose estima-tion network without refinement, and compare it with state-of-the-art methods that do not use refinement. Results in Table \ref{table:PPN} on three benchmarks suggest that PPN alone performs better than HMap and PoseCNN on all three datasets, and performs comparably to PVNet. 

Unlike these approaches, PPN has the highest speed (50 fps), is completely end-to-end, and does not require any additional steps such as the P$n$P algorithm. Thus, we suggest that PPN alone is fast and robust enough to be deployed in real-world applications.

\setlength{\tabcolsep}{4pt}
\begin{table}
\scriptsize
\begin{center}
\caption{Evaluation Results of our PPN compared to other state-of-the-art RGB-based methods that do not use refinement on three datasets: YCB-Video, LINEMOD and Occlusion using the 2D-Proj metric}
\label{table:PPN}
\begin{tabular}{|c|cccc|}
\hline 
Methods & PoseCNN\cite{xiang2017posecnn} & HMap\cite{oberweger2018making} & PVNet\cite{peng2019pvnet} & PPN(ours) \\
\hline
YCB-Video&3.72&39.4&47.4&\textbf{49.3}\\
\hline
LINEMOD&62.7&-&\textbf{99.0}& 96.12\\
\hline
Occlusion&17.2&60.9&61.06&\textbf{61.10}\\
\hline
\end{tabular}
\end{center}
\end{table}
\setlength{\tabcolsep}{1.4pt}

\section{Additional Qualitative Results}

\begin{figure}
\centering
\includegraphics[height=15cm]{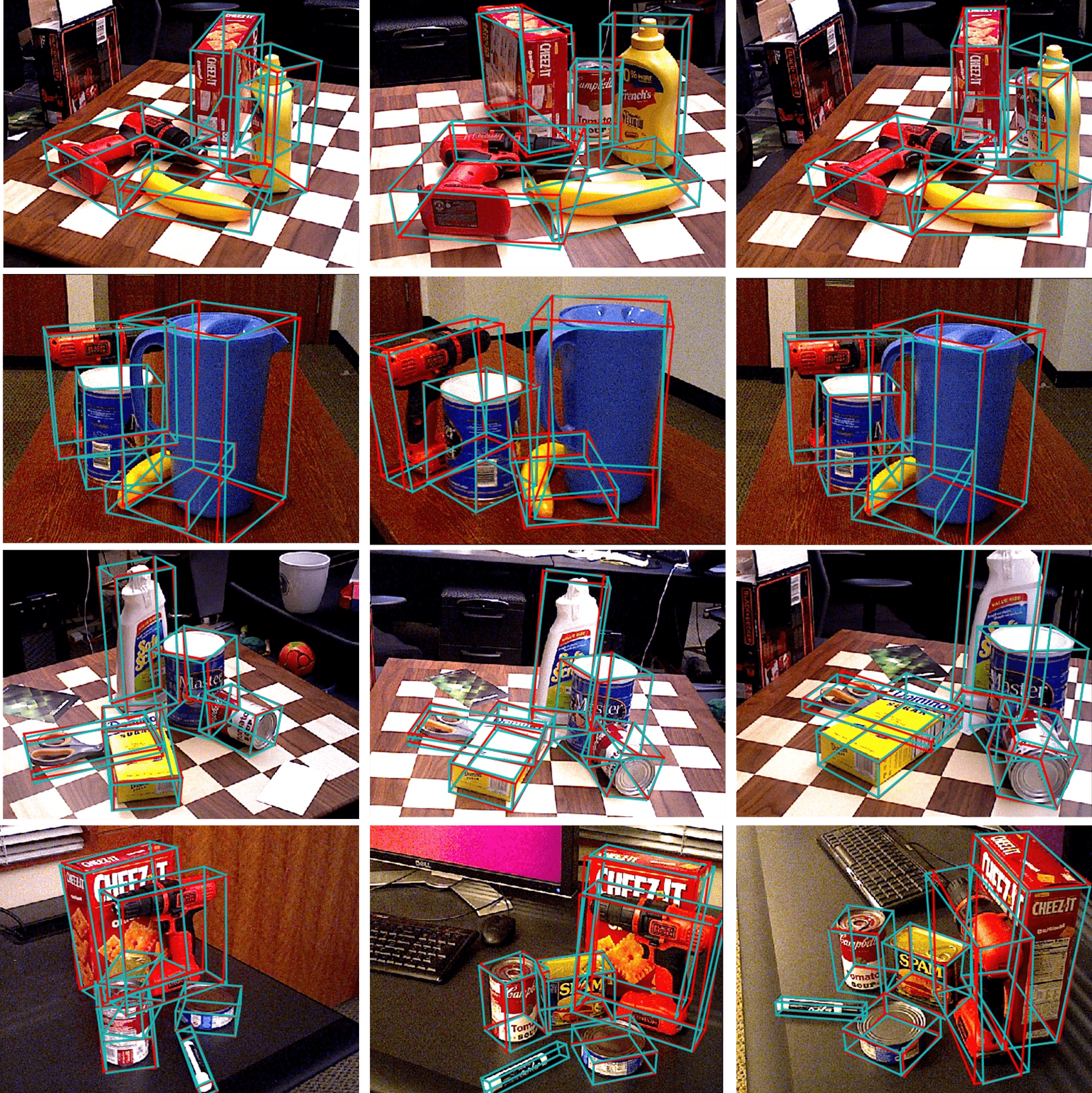}
\caption{Examples of 6D object pose estimation results on the YCB-Video dataset. Each row corresponds to images from one testing video. Red
bounding boxes correspond to ground truth poses, cyan bounding boxes correspond to predicted poses using our approach.}
\label{fig:rslts_qualt1}
\end{figure}

\begin{figure}
\centering
\includegraphics[height=14cm]{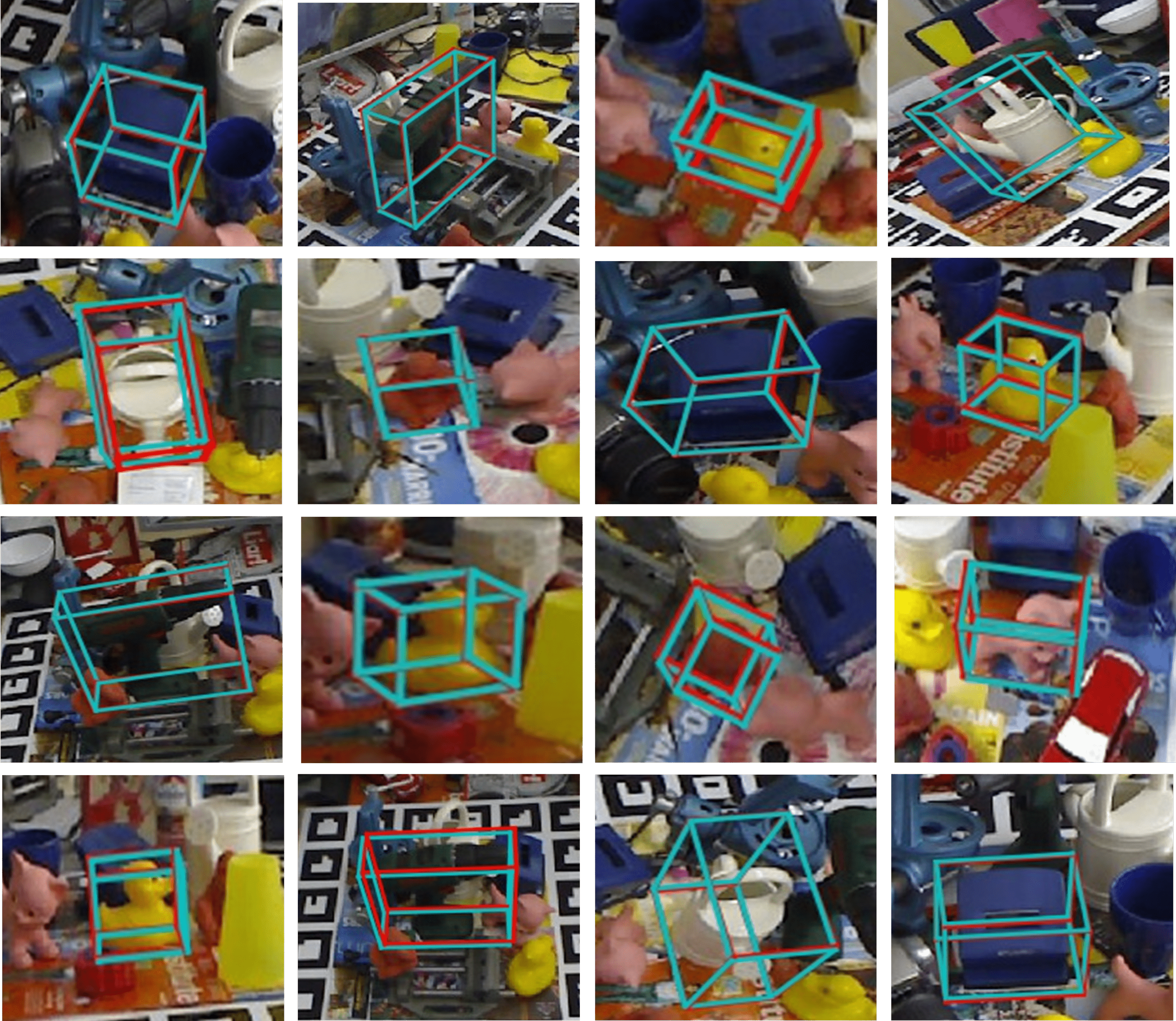}
\caption{Examples of 6D object pose estimation results on different objects from the LINEMOD dataset. Objects are: Holepuncher, driller, duck, can, ape, cat. Red
bounding boxes correspond to ground truth poses, cyan bounding boxes correspond to predicted poses using our approach.}
\label{fig:rslts_qualt2}
\end{figure}

\begin{figure}
\centering
\includegraphics[height=15cm]{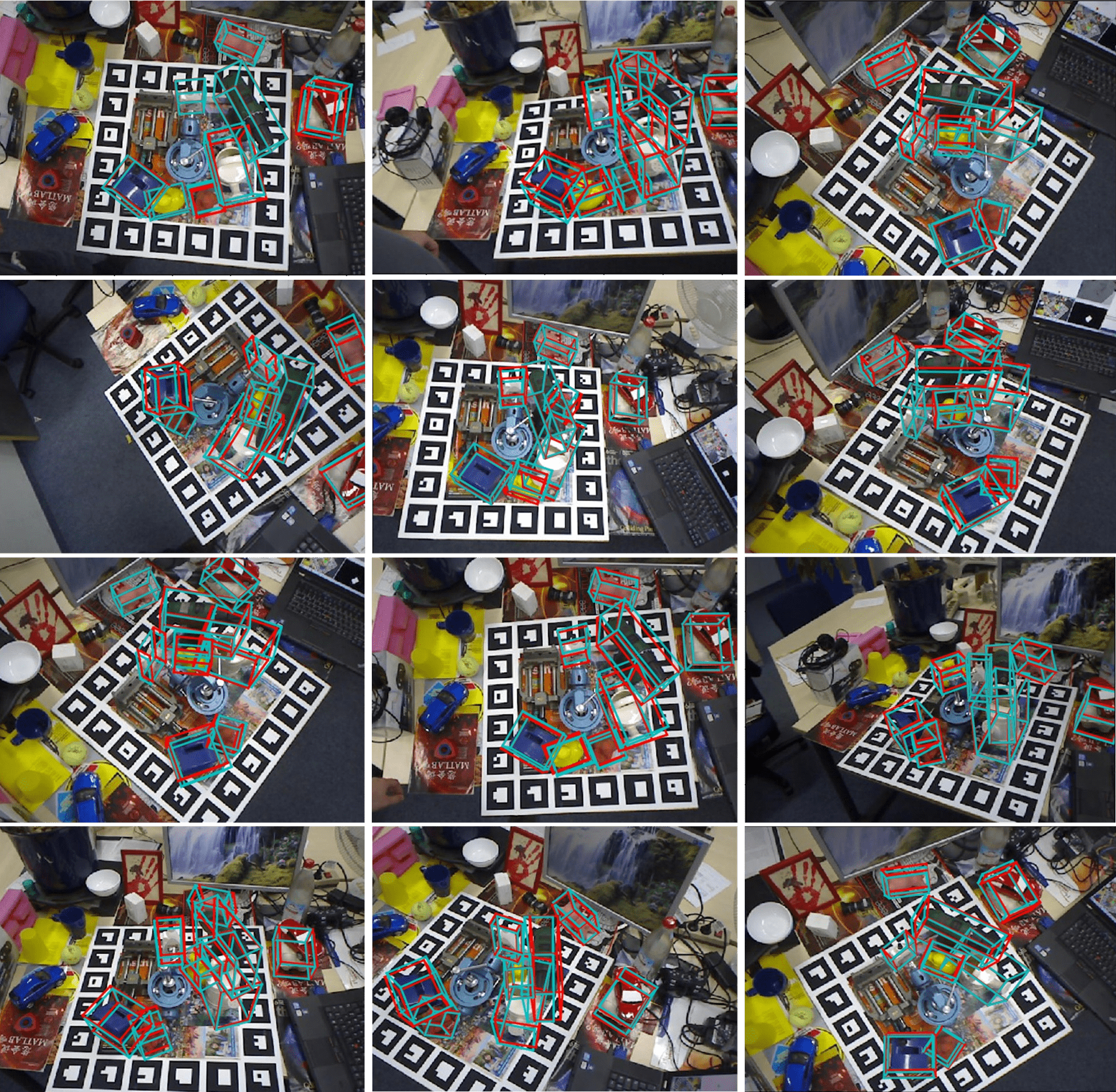}
\caption{Examples of 6D object pose estimation results on different objects from the Occlusion dataset. Red
bounding boxes correspond to ground truth poses, cyan bounding boxes correspond to predicted poses using our approach.}
\label{fig:rslts_qualt3}
\end{figure}

In \figurename~\ref{fig:rslts_qualt1} to \ref{fig:rslts_qualt3}, we show qualitative results on the three datasets: YCB-Video\cite{xiang2017posecnn}, LINEMOD\cite{hinterstoisser2012model} and Occlusion\cite{brachmann2014learning} datasets. These examples show that our proposed method is robust to severe occlusions, scene clutter, different illumination and reflection.

\section{Conclusion}
We have proposed a novel end-to-end method for RGB-only 6D pose estimation. Specifically, our end-to-end approach is mainly composed of two modules. First, PPN, is a fully-CNN-based architecture that produces a one-pass pose estimates. Second, MARN, is a pose refinement network that combines visual and flow features to estimate accurate transformations between the predicted and actual object pose. Further, MARN utilizes a spatial multi-attentional block to emphasize important feature parts, making the method more robust. Our full end-to-end model achieves state-of-the-art results on three separate datasets. 
\newpage

\bibliographystyle{unsrt}  
\bibliography{references}  

\end{document}